\documentclass{article}

\usepackage{arxiv}

\usepackage[utf8]{inputenc} % allow utf-8 input
\usepackage[T1]{fontenc}    % use 8-bit T1 fonts
\usepackage{hyperref}       % hyperlinks
\usepackage{url}            % simple URL typesetting
\usepackage{booktabs}       % professional-quality tables
\usepackage{amsfonts}       % blackboard math symbols
\usepackage{nicefrac}       % compact symbols for 1/2, etc.
\usepackage{microtype}      % microtypography
\usepackage{lipsum}		% Can be removed after putting your text content
\usepackage{graphicx}
\usepackage{subfig}
\graphicspath{images}
\usepackage{natbib}
\usepackage{doi}

\newcommand\freefootnote[1]{%
  \let\thefootnote\relax%
  \footnotetext{#1}%
  \let\thefootnote\svthefootnote%
}
\usepackage[hang,flushmargin]{footmisc}

\title{Learning-based Temporal Estimation of In-situ Wind Speed From Underwater Passive Acoustics}

\author{ Matteo Zambra\\
	IMT Atlantique, URM CNRS Lab-STICC\\
	Brest, France \\
	\texttt{matteo.zambra@imt-atlantique.fr} \\
	\And
	Dorian Cazau \\
	ENSTA Bretagne, URM CNRS Lab-STICC\\
	Brest, France \\
	\texttt{dorian.cazau@ensta-bretagne.fr} \\
	\And
	Nicolas Farrugia \\
	IMT Atlantique, URM CNRS Lab-STICC\\
	Brest, France \\
	\texttt{nicolas.farrugia@imt-atlantique.fr} \\
	\And
	Alexandre Gensse \\
	Naval Group\\
	Toulon, France \\
	\texttt{alexandre.gensse@naval-group.fr} \\
	\And
	Sara Pensieri \\
	IAS CNR Genova\\
	Genova, Italy \\
	\texttt{sara.pensieri@cnr.it} \\
	\And
	Roberto Bozzano \\
	IAS CNR Genova\\
	Genova, Italy \\
	\texttt{roberto.bozzano@cnr.it} \\
	\And
	Ronan Fablet \\
	IMT Atlantique, URM CNRS Lab-STICC \\
	Brest, France \\
	\texttt{ronan.fablet@imt-atlantique.fr}
	}
\date{}

\renewcommand{\shorttitle}{Learning-based estimation of in-situ wind speed from underwater acoustics}

\hypersetup{
pdftitle={zambra-et-al-2022},
pdfsubject={physics.comp-ph, physics.ao-ph, cs.LG},
pdfauthor={Matteo Zambra},
pdfkeywords={variational data assimilation, deep learning, geophysical dynamics, underwater acoustics, wind speed prediction},
}

\begin{document}
\maketitle

\begin{abstract}
Wind speed retrieval at sea surface is of primary importance for scientific and operational applications. Besides weather models, in-situ measurements and remote sensing technologies, especially satellite sensors, provide complementary means to monitor wind speed. As sea surface winds produce sounds that propagate underwater, underwater acoustics recordings can also deliver fine-grained wind-related information. Whereas model-driven schemes, especially data assimilation approaches, are the state-of-the-art schemes to address inverse problems in geoscience, machine learning techniques become more and more appealing to fully exploit the potential of observation datasets. Here, we introduce a deep learning approach for the retrieval of wind speed time series from underwater acoustics possibly complemented by other data sources such as weather model reanalyses. 
Our approach bridges data assimilation and learning-based  frameworks to benefit both from prior physical knowledge and computational efficiency. 
Numerical experiments on real data demonstrate that we outperform the state-of-the-art data-driven methods with a relative gain up to 16\% in terms of RMSE. Interestingly, these results support the relevance of the time dynamics of underwater acoustic data to better inform the time evolution of wind speed. They also show that multimodal data, here underwater acoustics data combined with ECMWF reanalysis data, may further improve the reconstruction performance, including the robustness with respect to missing underwater acoustics data. \freefootnote{\small  This work has been submitted to the IEEE for possible publication. Copyright may be transferred without notice, after which this version may no longer be accessible.}
\end{abstract}

% keywords can be removed

\section{Introduction}
Wind speed monitoring is of key interest for a wide range of applications and domains including climate and atmosphere science, meteorological modeling and routing applications. Besides in-situ measurements, remote sensing technologies have long been explored for the estimation of wind speed (\cite{li2020, Hersbach2010, RANA2019, jang2019}), in particular through satellite sensors. This class of techniques allows for a high-resolution and weather-independent sensing of the sea surface (like in the case of SAR imagery—\cite{wang2019}). However, the revisit time of satellite SAR sensors may be as large as 10 days, which cannot inform rapidly-evolving wind speed patterns from the hourly scale to the daily one.
This has motivated the exploration of other remote sensing technologies such as underwater acoustic sensors (\cite{zuba2015, nystuen2015}) to deliver complementary low-cost and non-intrusive indirect measurements of the sea surface wind speed. Such sensors can be deployed for a relatively long period of time, typically from weeks to years and can sample underwater noise continuously with a high time resolution.   

The seminal work of~\cite{nystuen1986} prompted the development of acoustic meteorology, a discipline that aims to reconstruct above-surface meteorological phenomena (such as wind speed and rainfall) given ocean ambient noise. The instruments typically used to capture underwater ambient noise are fixed hydrophones (\cite{ma2007, nystuen2010}), mobile acoustic platforms as ARGO profilers (\cite{yang2015, jie206}) and finally free-ranging bio-logged marine mammals (\cite{cazau2017}).
% The rest of this piece is mine
The developments of acoustic meteorology made possible to move towards an operationalization of passive acoustic measurements for geophysical phenomena, for example the estimation of wind speed starting from underwater noise measurement (\cite{nystuen2015, Cazau2018, asher2016}). Recently, machine learning techniques led to the further improvement of the prediction performance, as in \cite{taylor2020}. The appeal of machine learning approaches stems from their computational efficiency, the availability of open source libraries and the ability to skip handcrafted feature engineering steps.

From a methodological point of view, wind speed retrieval from acoustic data can be regarded as an inverse problem (\cite{snieder1999inverse}) which classically relies on state-space formulation when considering time-evolving processes. The complexity of the physical laws regulating the wind speed dynamics and the variability of the water column conditions that affects the underwater sound propagation, however, prevents the design of a classic model-driven inversion scheme. It also motivates the exploration of purely data-driven schemes in previous work by \cite{taylor2020, karpatne2018machine, bergen2019}, which do not however explicitly account for time-related patterns. Interestingly, recent studies have introduced physics-informed end-to-end learning schemes for inverse problems with time processes (\cite{fablet2021endtoend}). Here, we extend this line of work to the prediction of sea surface wind speed from underwater acoustics measurements. We exploit the 4DVarNet end-to-end architecture introduced by \cite{fablet2021learning} which relies on a variational data assimilation formulation. More specifically, it
represents the underlying physics through trainable neural network modules and addresses the associated optimization procedure through a trainable solver. 
While making explicit the modeling and inversion of time processes, we can learn all trainable components from data so that the underlying physical representation is optimized for the considered case-study. 4DVarNet is particularly suited for this application, given the time-dependent state-space statement of the underlying physical problem.
We also explore a multi-modal machine learning approach by incorporating different data sources, namely underwater acoustics and reanalyzed wind speed values available in the ERA-interim database (\cite{dee2011}). Multi-modality allows to incorporate heterogeneous information from different data sources (\cite{baltruvsaitis2018multimodal}). 

The remainder of the paper is organized as follows. Section~\ref{sec:problem-statement} introduces the relevance of the time dependence of wind speed and provides the suitable inverse problem formulation thanks to the state-space representation (\cite{hangos2004analysis}). Section~\ref{sec:dataset} presents our dataset and the multi-modal approach for this case study. Section~\ref{sec:methods} details the 4DVarNet end-to-end architecture and how it is applied to the present case study. Section~\ref{sec:results} presents the results obtained and section ~\ref{sec:conclusion} critically discusses these results.

\section{Problem statement}\label{sec:problem-statement}
State-of-the-art methods for the retrieval of sea surface wind speed from underwater acoustics data rely on machine learning schemes—see \cite{taylor2020}. They state the reconstruction problem as a regression problem, formally
\begin{equation}\label{eq:regression-plain}
    y_t \mapsto u_t
\end{equation}
% INSERT here an equation to state a time-dependent regression model
where $y_t$ refers to the acoustic spectrum at time $t$ and $u_t$ is the associated in-situ wind speed.

By contrast, the resolution of inversion problems in geoscience for time-related processes generally relies on a data assimilation formulation (\cite{evensen2009, bannister2017, blum2009, blayo2014advanced}). It explicitly accounts for the underlying temporal dynamics of the process of interest through a state-space model
\begin{equation}\label{eq:state-space-formulation}
	\left\{
		\begin{array}{r c l}
			\displaystyle \frac{\partial x (t)}{\partial t} &=& \mathcal{M}(x, t) \\[3mm]
			y(t) &=& \mathcal{H}(x, t)
		\end{array}
	\right.
\end{equation}
where $x \in X$ represents the state variable defined on the support $\left[ 0, T \right]$ and $\mathcal{M}$ is a given mathematical model of the physical dynamics. The variable $y \in Y$ represents a measurement of $x$ through observation operator $\mathcal{H}$. When dealing with irregularly-sampled observation data, the observation operator $\mathcal{H}$ involves a masking operator.
Besides dynamical model $\mathcal{M}$, we can define the associated flow operator $\Phi$  as the one-step-ahead predictor of the state $x(t+ \Delta t)$ based on the integration of the model $\mathcal{M}$ from time $t$
\begin{equation}
	\Phi(x, t + \Delta t) = x(t) + \int_{t}^{t + \Delta t} \mathcal{M}(x, t) \, dt
\end{equation}
The numerical implementation of flow operator $\Phi$ involves a numerical integration scheme such as Euler and Runge-Kutta explicit integration schemes (\cite{runge1924, cash2003}) using time discretization. Given the above state-space formulation, we can formulate a data assimilation problem to reconstruct state sequence $x$ from observation data $y$ as the minimization of the following variational criterion
\begin{equation}
	U_{\Phi}(x, y, \Omega) = \lambda_1 \, \sum_{t = 0}^{T} \| x(t) - y(t) \|^2_{\Omega} + \lambda_2 \, \sum_{t = 0}^{T} \| x(t) - \Phi(x; t) \|^2
\end{equation}
where $\Omega$ is the spatio-temporal domain where observations are defined. $\lambda_1$ and $\lambda_2$ are tunable weights. To state the previous equation more compactly we may use the vector notation
\begin{equation}\label{eq:variational-cost}
    U_{\Phi}(x, y, \Omega) = \lambda_1 \, \| x - y \|^2_{\Omega} + \lambda_2 \, \| x - \Phi(x) \|^2
\end{equation}
Then the optimization problems reads as follows:
\begin{equation}\label{eq:vc-optimization-statement}
    \tilde{x} = \displaystyle \arg \min_{x \in X} \; U_{\Phi}(x, y, \Omega)
\end{equation}
We may emphasize that this variational cost comprises two terms: a data fidelity term to assess the agreement between the reconstructed states and the observation data and a prior term to constrain the reconstructed states to match the dynamics associated with model $\mathcal{M}$. While a broad literature describes data assimilation algorithms to solve for the above minimization problem, a key issue for an application to in-situ wind speed retrieval is the ability to define and parameterize the underlying state-space formulation. Here, we may emphasize that physical models for the underwater acoustics dynamics as well as the relationship between underwater acoustics data and wind speed (\cite{Cazau2018, Pensieri2015}) appear too complex to explore a fully model-driven formulation. 

Interestingly, recent studies have explored how to bridge deep learning schemes and data assimilation methods (\cite{beauchamp2020, Ouala2020, nguyen2020assimilation, ouala2019learning, brajard2020combining, bocquet2020}) as a way to combine physically-sound formulations with the computational efficiency and the versatility of deep learning frameworks. As reported by \cite{fablet2021learning}, we have a generic end-to-end deep learning scheme for time-related inverse problems which explicit relies on a variational data assimilation formulation.

\begin{figure}
	\centering
	\subfloat{\includegraphics[width=0.325\linewidth]{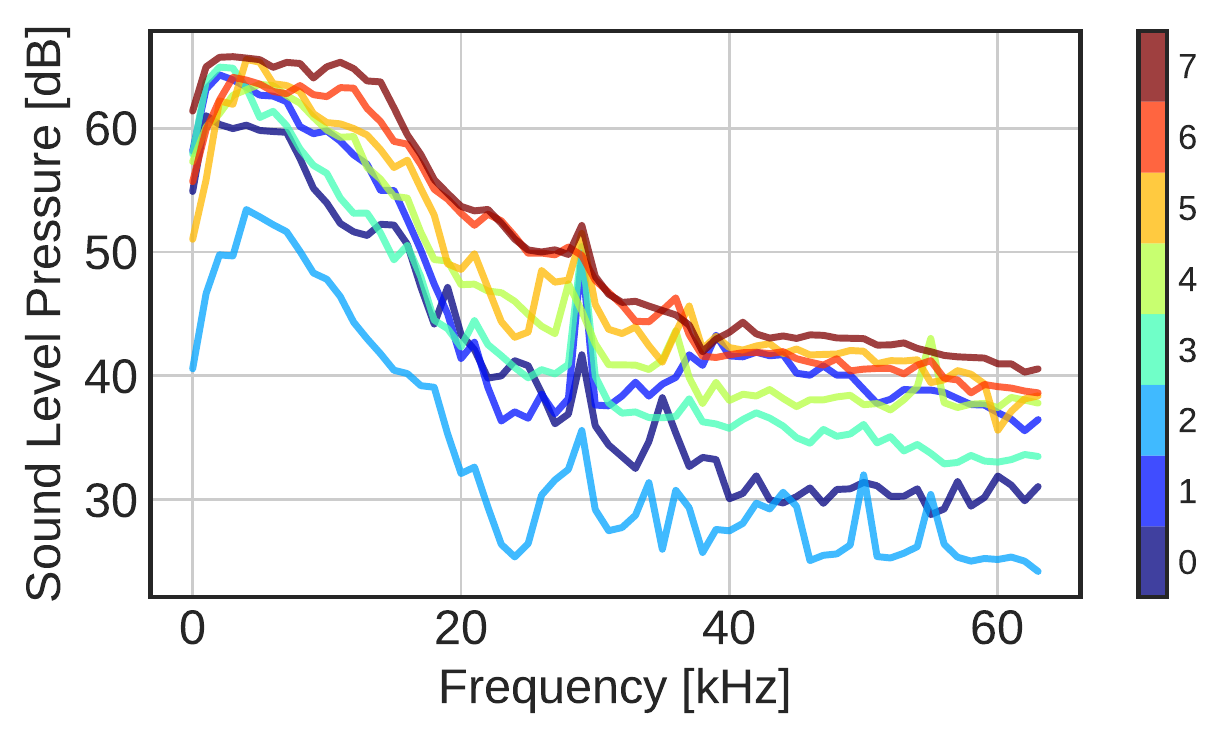}}%
	\quad
	\subfloat{\includegraphics[width=0.25\linewidth]{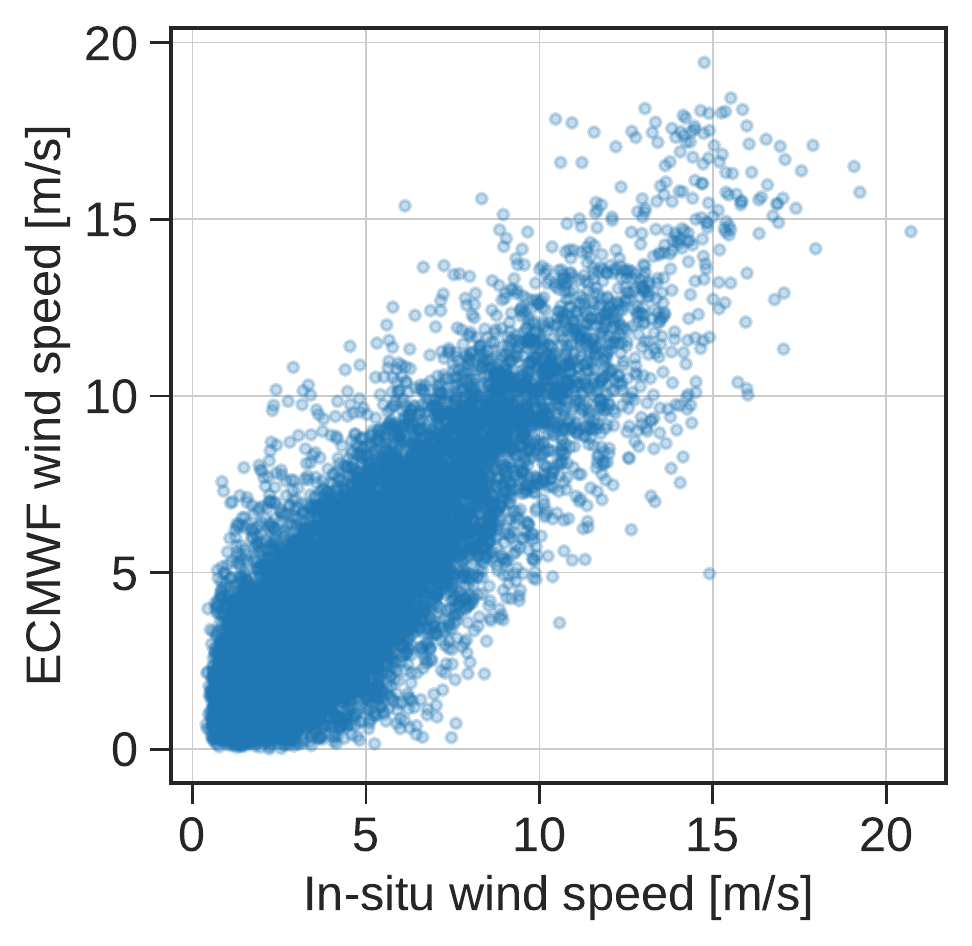}}%
	\quad
	\subfloat{\includegraphics[width=0.325\linewidth]{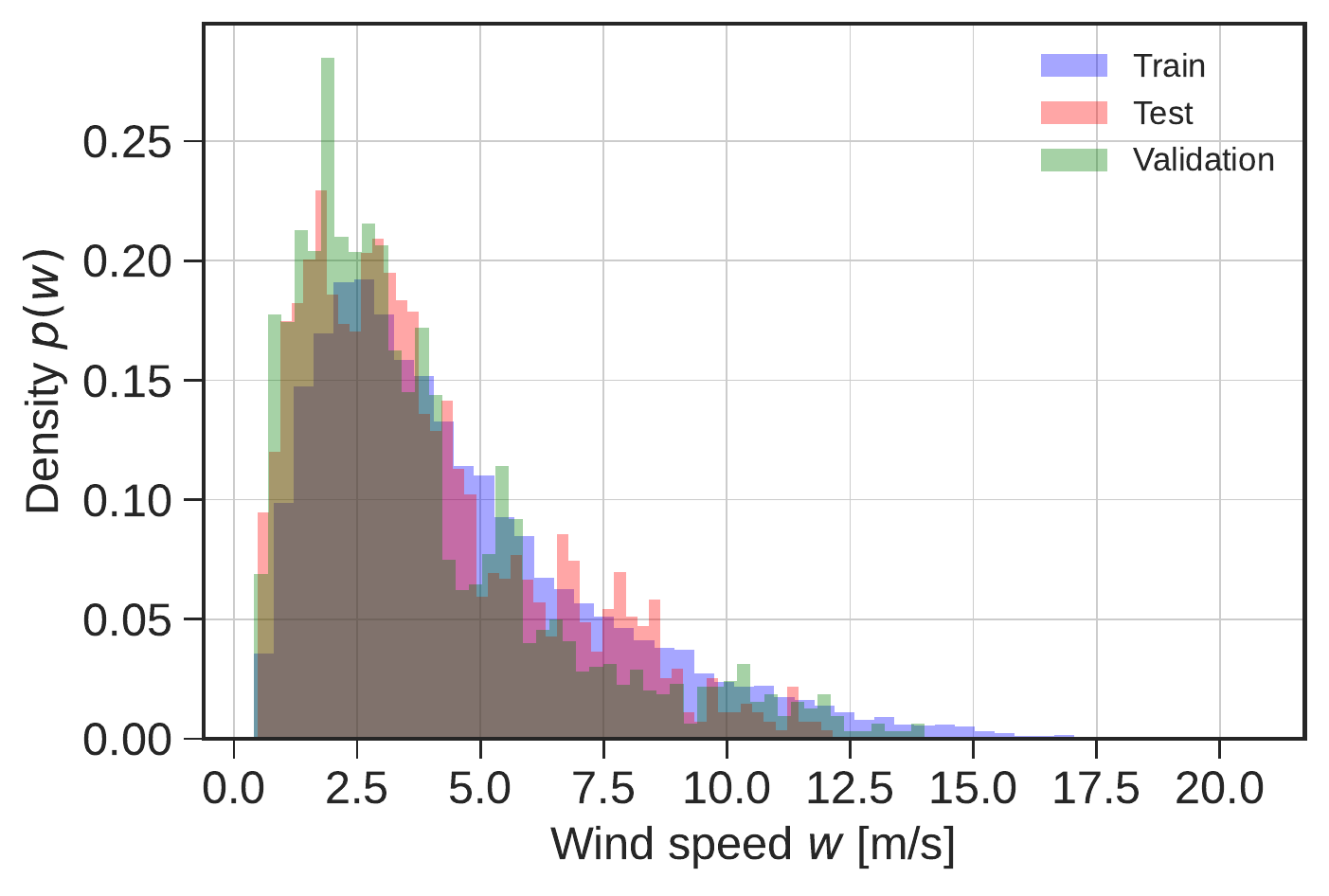}}%
	
	\caption{Data set quantitative and qualitative characteristics. Left panel: Different UPA spectra according to the associate in-situ wind speed, in Beaufort scale. Center panel: Scatter plot between in-situ and ECMWF wind speed. Right panel: Probability densities of in-situ wind speed. This is to quantify the values involved.}
	\label{fig:dataset}
\end{figure}

\section{Data}\label{sec:dataset}
Data used in this case study are underwater ambient noise spectra, synthetic reanalyses of wind speed provided by the European Center of Medium-Range Weather Forecast (ECMWF) and in-situ measurements of wind speed at 10 meters above the sea surface. ECMWF reanalyses are publicly available in the ERA-interim database (\cite{berrisford2011}) and, underwater passive acoustic (UPA) data and in-situ wind speed are sampled on the W1M3A marine observatory. A brief explanation on the physical functioning principles of the Underwater Passive Acoustic listener is provided in the subsection devoted to W1M3A. Figure~\ref{fig:dataset} gives a visual intuition of the data involved. Note that in-situ and ECMWF wind speed values are mildly correlated, $R^2 = 0.71$ and the root mean squared error between them is $1.71 \; m/s$. The in-situ wind speed distributions are displayed in the bottom right panel and show an high density on mild wind values. Higher grades of the Beaufort scale, up to grade 8\textsuperscript{th} are occasionally attained with a wind speed value of $20.71 \; m/s$.

\subsection{ECMWF wind speed values}
ECMWF wind speeds come from model reanalyses\footnote{In the data assimilation literature, a ``reanalysis'' is the reconstruction of given process over a given time horizon from a series of observation over the same  period of time based on the assimilation these observation data sources in a dynamical model.}.  These reanalyses are obtained by combining the available observations and the prior knowledge of the physical model (\cite{dee2011}). Since ECMWF wind is an estimation obtained with a numerical model, it is smoothed but implicitly carries information about the physical evolution of the meteorological variables involved. The wind speed reanalyses come from the ERA-Interim dataset, based on the global atmosphere model reanalysis developed at the ECMWF. All these global reanalyses are obtained with the assimilation of a large body of different in situ and satellite data. The atmospheric model is coupled to an ocean-wave model with a $1.0^{\circ} \times 1.0^{\circ}$ latitude and longitude grid. A detailed description of the ERA-Interim product archive can be found in the works by \cite{berrisford2011, dee2011}.

\subsection{The W1M3A observation system}
The observation site is the Western 1-Mediterranean Moored Multisensor Array (W1M3A) which is part of EMSO and ICOS networks of European research infrastructures. It is located in the Gulf of Genoa (Italy) at a distance of 80 km from the coast. The W1M3A system is composed of the ODAS Italia 1 spar buoy and a subsurface moored component. \cite{Pensieri2015} provide a detailed explanation of the W1M3A observation site.
Underwater noise data are collected with an Underwater Passive Acoustic Listener (UPAL). A detailed description of the functioning of UPAL are provided in \cite{zuba2015}, \cite{nystuen2015} and \cite{Pensieri2015}. For the sake of context, we provide here some insights of UPA data collection and preprocessing. UPAL is equipped with an hydrophone but, due to hardware constraints (battery life and memory) it is not possible to sample noise continuously during the whole instrument duty cycle. Noise is sampled for 4.5 seconds at 100 kHz and then processed to obtain a spectral representation of the signal having 64 frequency bands. The data used in this case study are indeed time series of such spectra. UPAL electronics is endowed with a recognition algorithm that classifies ambient noise sources. The duty cycle of the instrument is slightly adapted based on such rough classification, and on average, a recording is acquired every 5 minute. The frequency bands that better allow to recognise wind are those of 8 and 20 kHz, in fact more simple classification models based on UPA only account for these frequency bands. In our case, we leverage the capability of deep networks to minimize feature engineering and then we keep the whole spectrum. The interested reader may refer to \cite{nystuen2015} for a comprehensive explanation on the UPAL functioning.

In-situ wind speed is measured by means of the WindSonic-2D anemometer. This instrument measures the two horizontal components of the wind speed vector and thus the scalar speed modulus. 

\subsection{Temporal resolutions}
The underwater acoustics data are sampled from 2011-06-17 at 00:50 to 2013-09-06 at 18:50 almost continuously with an hourly resolution, except a period of time between 2013-04-26 and 2013-06-06 in which no observations are available. In-situ wind speed data are available from 2011-06-17 at 00:50 to 2013-09-06 at 18:50, except for a time window from 2012-11-06 to 2013-06-06 where no data are available. In-situ wind speed data are provided as the hourly average of the monitored wind speed values. In an open sea context, such as the W1M3A system, there is almost always non-negligible wind speed, except for some short periods of time (\cite{Pensieri2015}). Both UPA and in-situ wind data have hourly and co-located time resolution due to the fact that both data sources are hourly averages. ECMWF wind speed values are available from 2011-06-01 00:00 to 2019-06-30 23:00. Note that this latter data modality does not suffer from the presence of missing data, unlike UPAL and in-situ wind, since it is the output of an operational model. UPAL and in-situ wind speed time series, on the other hand, may have one or more missing time steps of observation, due to many reasons, including among others instrument failure or maintenance operations.

\subsection{Pre-processing scheme}
Our first pre-processing step consists in co-locating UPA, ECMWF and in-situ wind time series, according to their respective time resolutions. If one time step has only UPA and/or ECMWF but has not an in-situ wind speed value associated, this time step is removed from the overall dataset. If one time step has no UPA and/or ECMWF but has the in-situ value, it is kept since it may be a proof of robustness in case of missing data in our time series. To complete the data clean-up, we remove all the observation days that do not have full time series of 24 in-situ wind speed values.
This results in keeping about 98\% of the original dataset.   
In the subsequent analysis, we consider two 
different data structures. On the one hand, 
we explore each hourly sample individually within a time-independent scheme, which results in a dataset of 14088 hourly samples. On the other hand, we also consider a time-dependent analysis where the dataset is organized as a collection of 2000 sequences of 24 consecutive hourly samples for the training set. Note that if we simply aggregated time series depending on the date, then the overall train set would be composed of $14088 / 24 = 587$ time series. Rather, we randomly sample 24-hour sequences, even if some portions of these are overlapping.

\section{Proposed method}\label{sec:methods}
This section details the proposed end-to-end deep learning scheme based on a variational data assimilation formulation for the retrieval of in-situ wind speed from multi-source data, namely underwater acoustics and ECMWF data.

%% proposed section organization
\subsection{Proposed variational data assimilation model}\label{sec:proposed-method}
In section~\ref{sec:problem-statement} we introduced the formal statement of the problem. %Here a specification of the model in this case study is given.
Recall that in a variational data assimilation problem, one has access to some observations $y$ and wants to reconstruct their state variable $x$. Here, the state variable is assumed to be the concatenation of the different available data modalities. We consider in the following the outputs of single-modal and multi-modal versions of the proposed 4DVarNet framework. Assume that $\alpha$ indicates the UPA or UPA with ECMWF modality and $\beta$ the in-situ wind speed modality. In the single-modal version, the state variable is the concatenation of UPA and in-situ wind speed, $x \in \mathbb{R}^{N_{\alpha} + N_{\beta}}$, and the observable part is represented by UPA data, $y \in \mathbb{R}^{N_{\alpha}}$ with $N_{\alpha} = 64$ and $N_{\beta} = 1$. In the multi-modal version, the observable part is composed by the UPA and ECMWF data, and the state variable is still considered to be the concatenation of UPA, ECMWF and in-situ wind speed, {\em i.e.}  with $N_{\alpha} = 65$ and $N_{\beta} = 1$. Given these definitions of the state variable and of the observations, we can consider the following observation operator $\mathcal{H}$  
\begin{equation}\label{eq:obs-state-variables}
	y = \left[ y_{\alpha}, 0 \cdot x_{\beta} \right] = \mathcal{H} \left( \left[ x_{\alpha}, x_{\beta} \right] \right)
\end{equation}
This observation operator simply states that  no direct observation of the in-situ wind speed is available. The dependence between the in situ wind speed and the observed variables derives from the parameterization of prior operator $\Phi$. Rather than exploring an explicit ODE-based parameterization as in model-driven data assimilation schemes (\cite{bannister2017}), we rely on neural auto-encoder architectures as in \cite{fablet2021endtoend, fablet2021learning}. In such architectures, the inputs and outputs share the same shape. Auto-encoder architectures (\cite{baldi12a}) have been widely used for denoising, reconstruction and simulation tasks, see \cite{hinton2006}. They generally exploit a latent lower-dimensional representation of the input data (\cite{hinton2006}). This property seems appealing here to enforce the assumption of some underlying latent space jointly encoding sea surface wind speed and available observation data. 

Here we chose to use a 1D convolutional auto-encoder architecture (Conv-AE) (\cite{masci2011stacked}). This is motivated by the fact that convolutional networks can leverage trainable convolution operators to model translation-invariant features throughout data examples.
Our encoder is composed of two 1D convolutional layers, having input-output shapes of $(N_{\alpha} + N_{\beta}, 128)$ and $(128, 20)$. After the first layer a Leaky Rectified Linear Unit non-linear activation function is placed, its negative slope is set to $10^{-1}$. The decoder has the same structure, but in reverse order and the same non-linear activation functions are used after the two layers. All convolution layers involve a zero-padding and a kernel dimension of 3. The number of channels, $128$ and $20$ respectively, was set empirically from cross-validation experiments.

\subsection{Associated trainable solver}\label{sec:proposed-solver}
Within the proposed approach, the reconstruction of the state variable relies on the minimization of variational cost, stated in Equation~(\ref{eq:vc-optimization-statement}). We solve this optimization problem using another neural network model included in the end-to-end architecture. This trainable gradient solver, referred to as $\Gamma$, exploits a convolutional Long-Short Term Memory (LSTM) network (\cite{hochreiter1997}), the latter being particularly suited for time series modelling (\cite{siami2019performance}) and optimizer learning (\cite{andrychowitz2016}). Overall, the end-to-end architecture implements a predefined number of iterations of the following iterative rule:
\begin{equation}\label{eq:state-update}
	\left\{
	\begin{array}{r c l}
		g^{k} &=& \nabla_x U_{\Phi}(x, y) \\[3mm]
		x^{k + 1} &=& \mathcal{L} ~ \Gamma \left( g^{k}, h^{k}, c^{k} \right)
	\end{array}
	\right.
\end{equation}
where $\mathcal{L}$ is a linear layer to map the latent state $h$ of the LSTM cell to the state space.
In our implementation, the LSTM cell dimension was set to $100$, again based on numerical experiments. We may emphasize that these latent states of the LSTM cell, namely $h$ and $c$, are referred to with the $k$ superscripts, which denotes the $k$-th the iteration of the gradient solver. A typical choice for the total number of iterations ranges from $5$ to $10$. This choice derives both the computational complexity of the underlying automatic differentiation (\cite{pytorch2017autodiff}) as well as the ability of this gradient-based iterative update to converge with only very few gradient-based steps.

\subsection{Learning scheme}\label{sec:proposed-learningscheme}
Overall, the training procedure involves two nested gradient descents: an inner minimization  of variational cost (\ref{eq:variational-cost}) w.r.t.  state variable $x$  and an outer minimization of the training loss w.r.t. model parameters, especially the parameters of  models $\Phi$ and $\Gamma$, $\Theta$. The objective function for the former has been already discussed. Let $\Psi_{\Theta}$ be the 4DVarNet end-to-end system and $\hat{x} = \Psi_{\Theta} \left( y \right)$ the reconstructed state variable. Then, the reconstruction error can be evaluated as
\begin{equation}\label{eq:loss-function}
	L(x, y) = \lambda_d \, \| y_{\alpha} - \hat{x}_{\alpha} \|^2_{\Omega^{\alpha}} + \lambda_p \, \| y_{\beta} - \hat{x}_{\beta} \|^2_{\Omega^{\beta}}
\end{equation}
In this equation, $\Omega_{\alpha}$ and $\Omega_{\beta}$ are domain masks that account for missing data. These masks may be simple binary matrices that represent the data sparsity pattern. Parameters optimization is achieved with two different instances of the Adam algorithm (\cite{Kingma2015adam}). For both the Conv-AE and the LSTM solver the learning rate is set to $10^{-3}$ and the weight decay to $10^{-5}$. The weights $\lambda_d$ and $\lambda_p$ are set to $0.5$ and $1.5$ for the Conv-AE and 4DVarNet experiments. These weights were tuned empirically. The weight on wind prediction is greater because it is the term associated with the task of major interest. The full end-to-end architecture $\Psi_{\Theta}$ is trained for $200$ epochs, with no early stopping criteria.
Our learning protocol consists of two consecutive steps. First we perform a full training procedure with 5 gradient iterations for the 4DVarNet scheme  and save the best model based on validation loss. Note the best model is not necessarily the one at the last epoch of the training procedure. Then, this best model is further trained through another full training procedure, this time using 10 gradient iterations.

\begin{figure*}
	\centering
	\subfloat{\includegraphics[width=\linewidth]{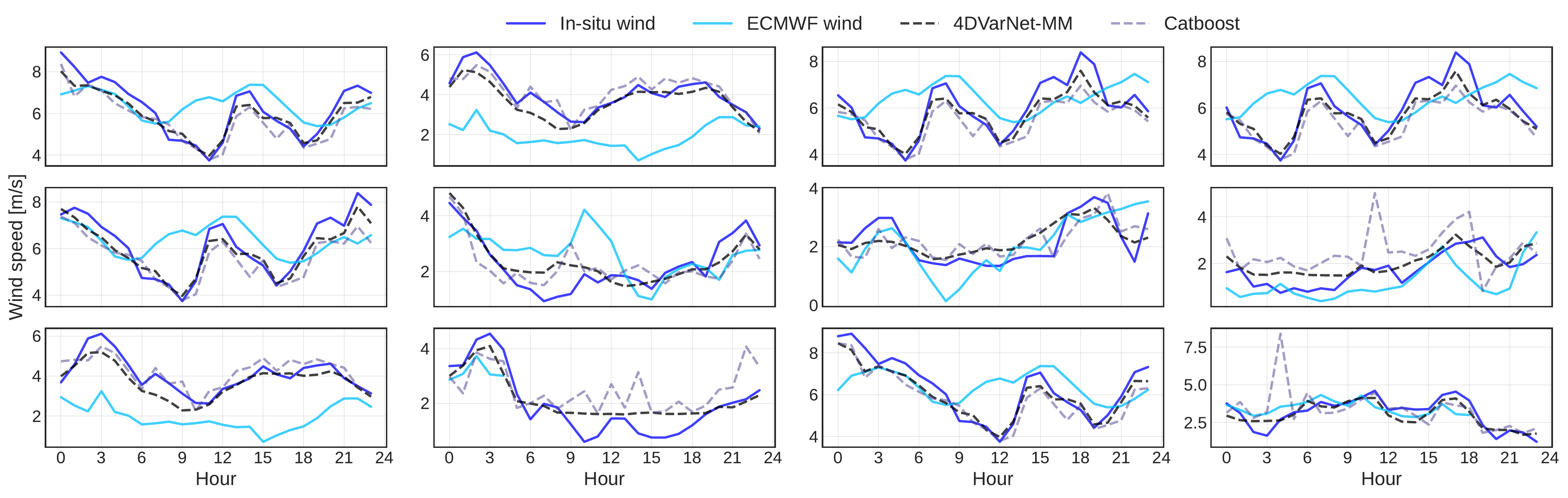}}%
	
	\caption{Wind speed reconstructions: examples of wind speed reconstruction over different 24-hour windows in the test dataset. We depict the in situ wind speed and compare the reconstructions issued from the proposed approach to Catboost reconstruction, \cite{taylor2020} and ECMWF wind speed.}
	\label{fig:wind-speed-comparison}
\end{figure*}

\section{Results}\label{sec:results}
This section reports the numerical experiments carried out to assess performance of the proposed approach with respect to state-of-the-art data-driven methods\footnote{We refer the reader to the following in the repository for our associated implementation \texttt{https://github.com/CIA-Oceanix/4DVarNet-wsp}}. We first detail our evaluation setting and the benchmarked models. We then report and discuss the reconstruction performance for three case studies for the reconstruction of in situ wind speed (i) using only underwater acoustics data, (ii) using underwater acoustics and ECMWF data, (iii) when dealing with random gaps in the underwater acoustics data, again in a multi-modal UPA and ECMWF dataset case.

\subsection{Evaluation framework}
Our evaluation procedure follows the one reported by \cite{taylor2020}. We chose as evaluation metrics the root mean squared error (RMSE) between reconstructed and in situ wind speed values. We evaluate this RMSE on data period from 2011-06-18 00:00 to 2011-08-07 23:00 (50 days) whereas data period 2011-08-08 00:00 to 2013-09-05 23:00 is used as an independent training dataset, except for the period between 2011-08-25 00:00 to 2011-10-14 23:00 (50 days) which is used as validation set. From the 50-day time series with an hourly time resolution, we extract all associated one-day time windows, which result in a dataset of $(50 - 1) \times 24 = 1176$ 24-hour samples.

For benchmarking purposes, our numerical experiments involve machine learning schemes proposed by \cite{taylor2020} as well as state-of-the-art neural network architectures. The later have been chosen in accordance with the parametrization of 4DVarNet scheme. Overall, the first category of methods involves machine learning schemes, referred to as time-independent, which predict in-situ wind speed at one time step from the underwater acoustics spectrum at the same time step, as stated in Equation~(\ref{eq:regression-plain}). The multi-modal approach is not implemented for this class of models. This category includes:
\begin{itemize}
    \item {\bf CatBoost} (\cite{Prokhorenkova2018}) A gradient-boosting algorithm (\cite{friedman2001greedy, friedman2002stochastic}) which can manage effectively categorical features. CatBoost uses as loss function the root mean square error as loss function.
    \item {\bf Random Forest} (\cite{Breiman2001}) An ensemble of decision trees, either trained for classification or regression. Random Forest has set up with the maximum depth of the tree to $10$, the number of features to consider when splitting equal to the actual number of features, the minimum number of samples required for leaf nodes to $1$ and the minimum number of samples required to split an internal node to $2$. The number of estimators is set to $100$.
    \item {\bf FC-AE}. This Fully-Connected Auto-Encoder (FC-AE) architecture comprises a fully-connected encoder composed of 2 linear layers with input and output shapes of $(N_{\alpha} + N_{\beta}, 128)$ and $(128, 20)$, respectively. After the first layer a Leaky Rectifier Linear Unit with negative slope of $0.1$ is applied. The decoder has the same architecture but reversed, with the same non-linearity applied after all layers. The learning rate is set to $10^{-3}$ and the weight decay is set to $10^{-6}$. The weights of the loss terms are chosen to be $0.5$ and $1.5$. The intermediate and latent dimensions, learning rate and weight decay and finally the loss terms weights were set empirically using cross-validation.
\end{itemize}
The second category of machine learning methods, referred to as time-dependent methods, predicts a time series of in-situ wind speed from a time series of underwater acoustics data and may benefit from time-related features. This category comprises the following schemes:
\begin{itemize}
    \item {\bf Fully-connected auto-encoder}. The FC-AE parametrization as described above has been reused in the time-dependent configuration. There are no differences in the architecture as the input size does not change, except the temporal dimension which is neglected in the time-independent setting.
    \item {\bf Conv-AE-UPA}. This convolutional auto-encoder architecture refers to the parameterization of operator $\Phi$ in Equation~(\ref{eq:variational-cost}) for the proposed 4DVarNet scheme. Here, we use this architecture to train a direct inversion scheme to map input data $y$ to a reconstructed state $x$. We may point out that this direct inversion scheme can be regarded as a single iteration of a fixed-point iterative solver for the minimization of variational cost (\ref{eq:variational-cost}).
    \item {\bf Conv-AE-UPA+ECMWF}. this architecture refers to an extension of the previous one when the input data $y$ includes both UPA and ECMWF wind speed. Besides, the reconstruction $x$ also comprises UPA, ECMWF and in-situ wind speed states.
    \item {\bf 4DVarNet-UPA}. Using the Conv-AE as previously described and the trainable solver as detailed in section~\ref{sec:proposed-solver}, the first 4DVarNet configuration accounts for the observations $y$ of UPA data only. The output $x$ is a concatenation of UPA reconstructions and in-situ wind speed predictions.
    \item {\bf 4DVarNet-UPA+ECMWF}. In this second case, the architecture of the 4DVarNet architecture is the same, except for the observations and the state variable. The input data comprehend both UPA and ECMWF wind speed.
\end{itemize}

The loss function used to train the deep models is a simple mean squared error, formulated in Equation~(\ref{eq:loss-function}). For evaluation on the test set, the root mean squared error is instead used. Since the task studied is the reconstruction of wind speed time series, we consider performance metrics based on the in-situ wind speed data considered as ground-truth.. To assess uncertainties and have reliable results, 10 runs are performed on a machine equipped with 3 Nvidia Quadro RTX 8000 units. Each of these units has a TU102 graphical processor operating at a frequency of 1395 MHz and has 48 GB memory size. In the following tables two columns are present. A first column, named ``Mean ± std'' reports the average RMSE and the quartiles over the 10 runs. In order to compare our framework with ensemble models previously discussed, another evaluation strategy is used. Each of the 10 models trained can produce a reconstruction of wind speed sequence given test data. Similarly to what is done in \emph{bagging} (\cite{Breiman1996}), we chose to compute the median of the wind speed values reconstruction as aggregated output. The $n$-Median metric is defined as the RMSE between this aggregated reconstruction and the ground truth wind speed values.

In order to quantify the improvement of the proposed class of models with respect to the baselines, we may define a relative gain metric. Call $p_b$ and $p_i$ the baseline and improved performance metrics, chosen to be the ensemble $n$-Median scores. Then define the relative gain as
\begin{equation}
    \eta = \left( 1 - \frac{p_i}{p_b} \right) \cdot 100
\end{equation}
In the following tables, the relative gain scores are reported for each of the associated models. In the results related to the time-independent models, such a comparison is not extremely informative. Rather in this first comparison we assess what is the most indicative baseline model to perform the subsequent comparisons.

\begin{table}
	\centering
	\caption{RMSE metrics for time-independent dataset and models.}
	\label{tab:ti-data-models}
	\begin{tabular}{l c c c}
	\hline
		Model & \multicolumn{3}{c}{Metrics in $m/s$} \\
		      & RMSE & Mean ± std & $n$-Median \\
		\hline
		CatBoost      & $0.95$ & -- & -- \\
		Random Forest & $0.97$ & -- & -- \\
		FC-AE         & -- & $0.98 \pm 0.03$ & $0.95$ \\
		\hline
	\end{tabular}
\end{table}

\subsection{UPA-only time-independent models}
The performance of time-independent models are reported in Table~\ref{tab:ti-data-models}. FC-AE performs as well as classical regression models. In a time-independent scenario, where the interest is only the prediction of a wind speed label given a single acoustic spectrum, a framework like the one described in Equation~(\ref{eq:regression-plain}) could then be preferable. Since in the time-independent configuration the fully connected auto-encoder and the CatBoost have a similar performance, see Table~\ref{tab:ti-data-models}, the performance metric of these models is now taken as baseline in order to evaluate the improvement hereafter. In this table, no data are provided for the mean and quartiles for CatBoost and Random Forest, since these are ensemble methods and hence they give yet an aggregated output.

\begin{table}
	\centering
	\caption{Results for single and multi-modal settings.}
	\label{tab:single-multi-modal-results}
	\begin{tabular}{l c c c}
		\hline
		Model & \multicolumn{2}{c}{Metrics in $m/s$} &  \\
		                           & Mean ± std & $n$-Median & $\eta$ [\%]\\
		\hline
		FC-AE 		               & $0.97 \pm 0.04$ & $0.92$ & $3.2$ \\
		Conv-AE-UPA                & $0.94 \pm 0.04$ & $0.88$ & $7.4$ \\
		4DVarNet-UPA 5 iter        & $0.91 \pm 0.02$ & $0.87$ & $8.4$ \\
		4DVarNet-UPA 10 iter       & $0.89 \pm 0.04$ & $0.84$ & $11.6$ \\
		Conv-AE-UPA+ECMWF          & $0.88 \pm 0.02$ & $0.83$ & $12.6$ \\
		4DVarNet-UPA+ECMWF 5 iter  & $0.85 \pm 0.03$ & $0.81$ & $14.7$ \\
		4DVarNet-UPA+ECMWF 10 iter & \boldmath{$0.84 \pm 0.02$} & \boldmath{$0.80$} & \boldmath{$15.8$} \\
		\hline
		\vspace{0.25mm}
	\end{tabular}
\end{table}

\subsection{UPA-only time-dependent case}
Table~\ref{tab:single-multi-modal-results} displays the results given by the model detailed in section~\ref{sec:methods}. The case of single-modal dataset presents a relative improvement of the performance with respect to the time-independent configurations. FC-AE applied in a time-dependent scenario has a similar performance as in the time-independent case. Indeed the Conv-AE model yet suffices to improve the time-independent FC-AE baseline by $7.4 \%$. The 4DVarNet-UPA leads to a relative gain of $8.4 \%$ with respect to the FC-AE and regression models baselines.

\subsection{Multi-modal time-dependent case}
Results in Table~\ref{tab:single-multi-modal-results} confirm the potential of the multi-modal approach. The Conv-AE trained on a heterogeneous dataset gives a gain of $12.3 \%$ with respect to the FC-AE benchmark. The 4DVarNet-UPA+ECMWF model, the multi-modal counterpart of the single modal 4DVarNet-UPA, yields a performance gain of $14.7 \%$ and $15.8 \%$ with respect to the FC-AE benchmark and the state-of-the-art presented by previous work. Recall that such a result by the 4DVarNet was obtained through the training protocol explained in section~\ref{sec:proposed-learningscheme}: 5 gradient iterations on the first training step and 10 iterations on the second step after selection by best score on the validation set. Figure~\ref{fig:wind-speed-comparison} presents a visual comparison between ground truth wind speed time series and the reconstructions obtained with selected models. ECMWF wind values are also super-imposed. These time series are compared with the reconstruction performed by a time-independent and a time-dependent model, the CatBoost and the multi-modal 4DVarNet, respectively. 

\begin{figure}
	\centering
	\subfloat{\includegraphics[width=0.45\linewidth]{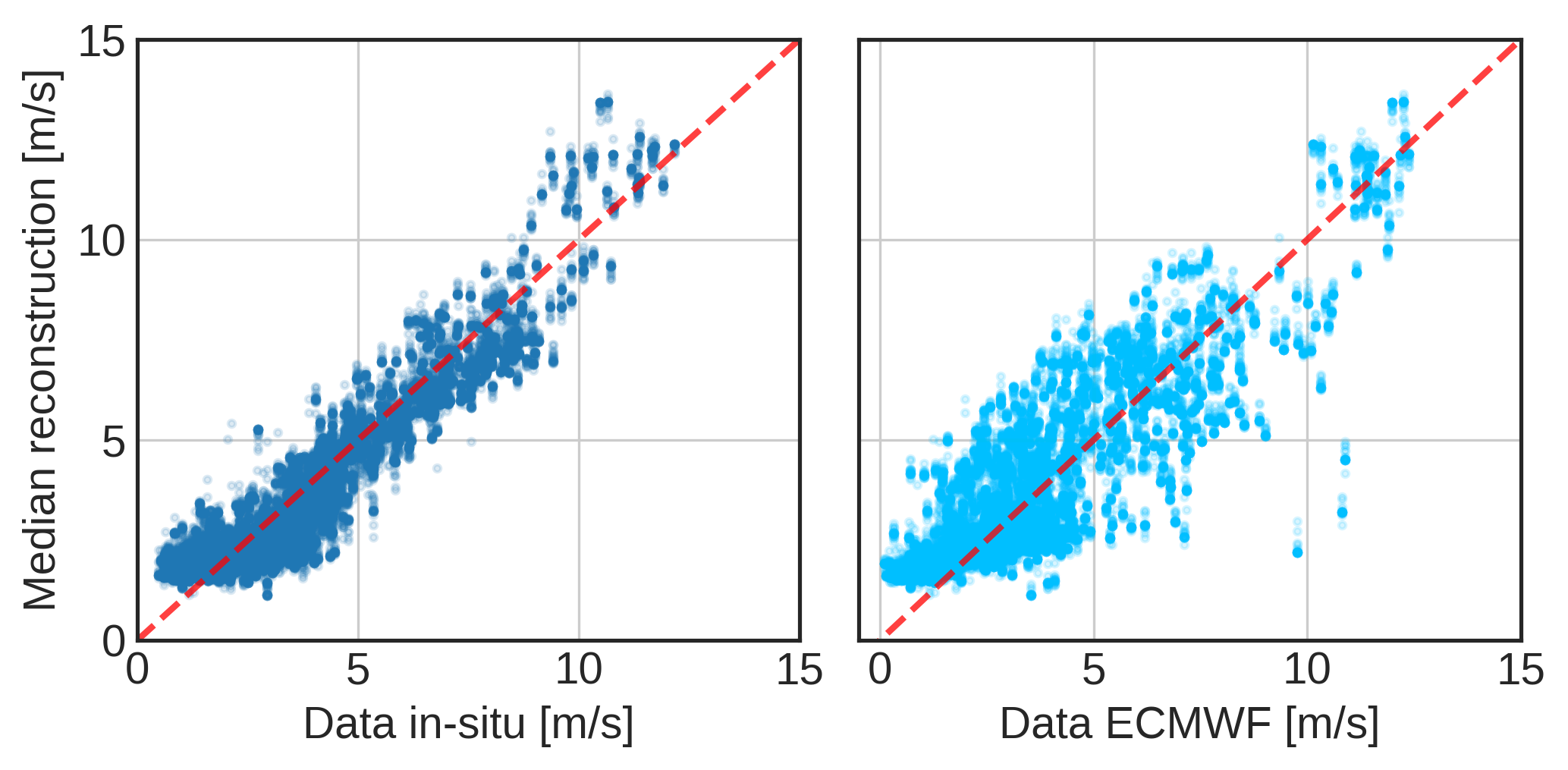}}%
	\quad
	\subfloat{\includegraphics[width=0.45\linewidth]{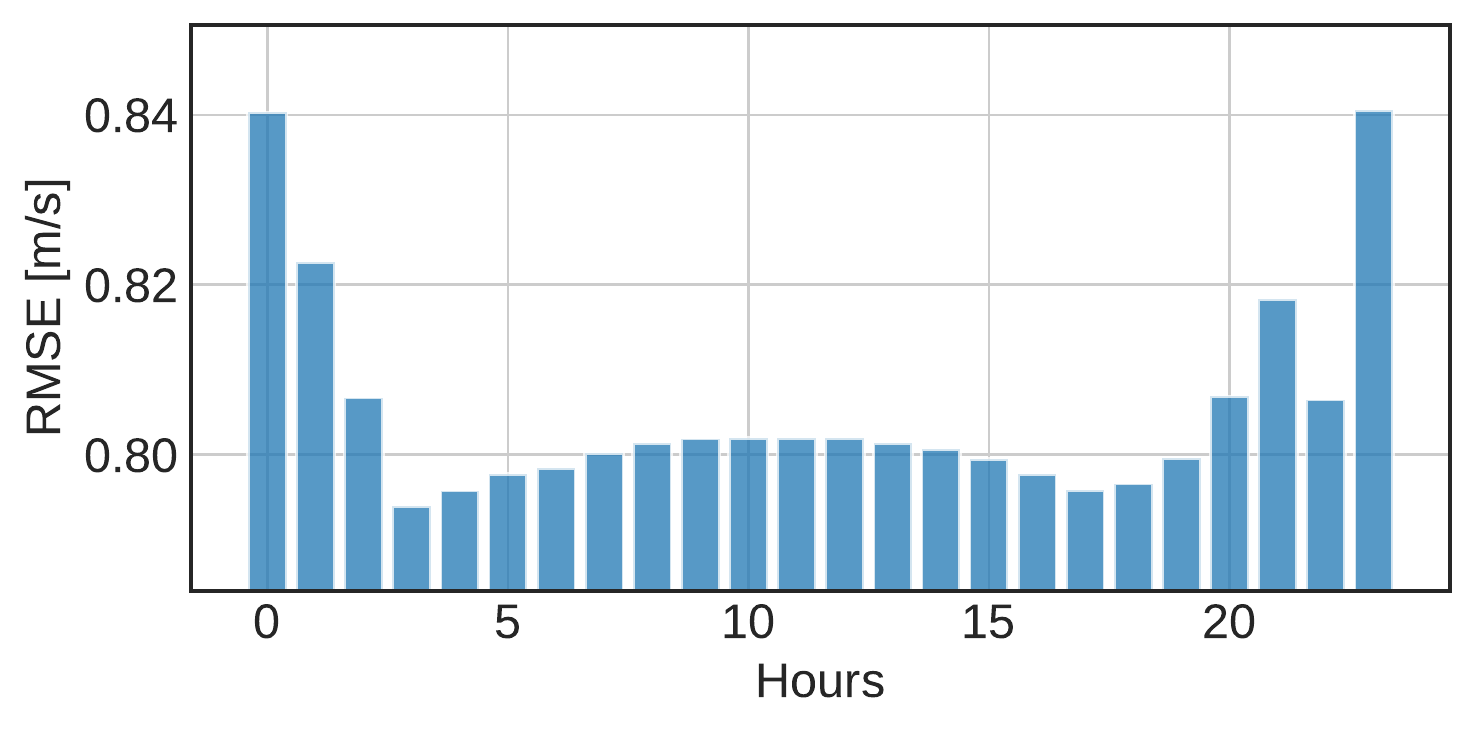}}%
	\caption{Overall reconstructions and error patterns. Top panel: Scatterplots of reconstructions of wind speed obtained with 4DVarNet against in-situ ground-truth and ECMWF wind speeds. Bottom panel: Hourly averages of reconstruction error.}
	\label{fig:reco-scatter-error-evaluation}
\end{figure}

Figure~\ref{fig:reco-scatter-error-evaluation} shows a scatterplot between the reconstructed and ground truth values. The top left panel clearly highlights a bias in the reconstruction of high wind speed values. This could be due to saturation of underwater acoustic data for large wind speeds. The bottom panel presents the average hourly error. This plot shows the effect of 1D convolutional filters, since the boundaries of the time observation intervals are not entirely involved by the striding of the filters hence there is no coverage backward and forward. One other interesting feature of this plot is the worse predictive performance on the central part of the day. This might be due to the diurnal cycle of the wind with the weakest wind speed during the day and strongest winds during the nights. Additionally, ECMWF modeled winds are also known to suffer from a bias usually ascribable to a misrepresentation of mesoscale convective variability and wind shear \cite{rivas2019}. 

\subsection{Multi-modal time-dependent case with missing data}
One of the points of interest of 4DVarNet is that it can handle time series with missing and/or corrupted data. Since underwater acoustic data in our dataset are almost complete, experiments on missing data were designed by arbitrarily artificially masking data batches during training and testing. The missing data percentage is a parameter chosen to range between $10 \%$ and $90 \%$. Table~\ref{tab:missing-data-results} reports the complete set of our experiments. Note that when $10 \%$ of the data is missing, 4DVarNet performs as good as the best multi-modal model trained with complete data. One may argue that removing $10 \%$ of data is analogous to dropout mechanics, artificially removing features and/or noising the data samples (\cite{wager2013, srivastava14a}). We may also note that the proposed approach reaches almost the same performance as the CatBoost model presented by \cite{taylor2020} up to $70\%$ of missing data.

\begin{table}
	\centering
	\caption{Missing data, 4DVarNet-UPA+ECMWF 10 iter.}
	\label{tab:missing-data-results}
	\begin{tabular}{c c c c}
		\hline
		Missing data \% & \multicolumn{2}{c}{Metrics in $m/s$} & \\
		& Mean ± std & $n$-Median & $\eta$ [\%] \\
		\hline
		$10$ & $0.85 \pm 0.02$ & $0.80$ & $15.8$ \\
		$20$ & $0.87 \pm 0.02$ & $0.81$ & $14.7$ \\
		$30$ & $0.90 \pm 0.02$ & $0.83$ & $12.6$ \\
		$40$ & $0.92 \pm 0.02$ & $0.83$ & $12.6$ \\
		$50$ & $0.98 \pm 0.02$ & $0.89$ & $6.3$ \\
		$60$ & $1.01 \pm 0.02$ & $0.91$ & $4.2$ \\
		$70$ & $1.07 \pm 0.03$ & $0.96$ & $-1.1$ \\
		$80$ & $1.17 \pm 0.03$ & $1.08$ & $-13.7$ \\
		$90$ & $1.27 \pm 0.03$ & $1.21$ & $-27.3$ \\
		\hline
	\end{tabular}
\end{table}

\section{Conclusion}\label{sec:conclusion}
This paper presented a novel robust and efficient framework for managing time dependence in data sets comprehending underwater acoustics and wind speed values. While previous work successfully demonstrated that machine learning approaches are promising tools to perform wind speed estimation given underwater acoustic data, this work highlights and proves the importance of explicitly accounting for time dependence. This concept could be further expanded to short then long-term forecasting problems, that is predicting a time series of wind speed given a small amount of acoustic data, in such a way to forecast the wind trend in a near future window. 

Further work may also consider the joint use of acoustic data and satellite imagery, such as Synthetic Aperture Radar (SAR) images. While acoustic data have a limited spatial coverage but have a rich resolution in time, SAR images display opposite characteristics, as they are scarce in time but offer a wider spatial resolution. A multi-modal approach that bridges these two temporally-rich and spatially-rich features could lead interesting research directions aiming to fit trainable models in which learning one modality helps in learning the other. Cross-modal learning and generation is a salient and important feature of multi-modal machine learning—see \cite{baltruvsaitis2018multimodal}. 

A further improvement could address the target variable to be modelled. In this case study, we mainly focused on the prediction of an environmental variable, but other important applications could benefit from the use of underwater ambient noise for anthropic activities such as submarine recognition or sea wildlife observation.

\section*{Acknowledgment}
This work is funded by the AI Chair Oceanix (ANR grant ANR-19-CHIA-0016) and is supported by the industrial partnership with Naval Group. 

%\bibliographystyle{unsrtnat}
%\bibliography{template}

\end{document}